\documentclass{article}
\usepackage{iclr2023_conference,times}

\usepackage{hyperref}
\usepackage{url}
\usepackage{graphicx}
\usepackage{amsmath}
\usepackage{booktabs}
\usepackage{color}
\usepackage{xcolor}
\usepackage{tikz}
\usepackage{enumitem}
\usepackage{fontawesome}
\usepackage{hyperref}
\usepackage{caption}
\usepackage{subcaption}
\usepackage{makecell}

\newcommand{\arnav}[1]{\textcolor{black}{#1}}
\newcommand{\olid}{{\textit{OLID}}}

\def\checkmark{\faCheck\space} 
\def\cross{\faClose\space}
\def\vague{\faQuestion\space}
\def\implicit{\faEyeSlash\space}

\title{Detecting Harmful Content on Online Platforms: What Platforms Need vs. Where Research Efforts Go}

\author{Arnav Arora\\
Checkstep Research;\\
University of Copenhagen\\
Copenhagen, Denmark\\
\texttt{aar@di.ku.dk}
\And
Preslav Nakov\\
Checkstep Research; \\
Mohamed bin Zayed University of Artificial Intelligence\\
Abu Dhabi, UAE
\And
Momchil Hardalov\\
Checkstep Research;\\ Sofia University ``St. Kliment Ohridski''\\
Sofia, Bulgaria
\And
Sheikh Muhammad Sarwar\\
Checkstep Research;\\ University of Massachusetts, Amherst\\
USA
\And
Vibha Nayak\\
Checkstep \\
London, UK
\And 
Yoan Dinkov\\
Checkstep Research\\
Sofia, Bulgaria
\And
Dimitrina Zlatkova\\
Checkstep Research\\
Sofia, Bulgaria
\And 
Kyle Dent\\
Checkstep\\
London, UK
\And
Ameya Bhatawdekar\\
Checkstep; Microsoft\\
Doimukh, Arunachal Pradesh
\And 
Guillaume Bouchard\\
Checkstep\\
London, UK
\And
Isabelle Augenstein\\
Checkstep Research; \\
University of Copenhagen\\
Copenhagen, Denmark
}

\iclrfinalcopy %
\begin{document}

\maketitle

\begin{abstract}
The proliferation of harmful content on online platforms is a major societal problem, which comes in many different forms including hate speech, offensive language, bullying and harassment, misinformation, spam, violence, graphic content, sexual abuse, self harm, and many other. Online platforms seek to moderate such content to limit societal harm, to comply with legislation, and to create a more inclusive environment for their users. Researchers have developed different methods for automatically detecting harmful content, often focusing on specific sub-problems or on narrow communities, as what is considered harmful often depends on the platform and on the context. We argue that there is currently a dichotomy between what types of harmful content online platforms seek to curb, and what research efforts there are to automatically detect such content. We thus survey existing methods as well as content moderation policies by online platforms in this light and we suggest directions for future work.
\end{abstract}

\section{Introduction}
\label{introduction}

Online harms is a serious and growing problem, and targeted groups and individuals 
have suffered it for years. 
There are various types of harms, ranging from clearly illegal activities (e.g.,~child sexual abuse, human trafficking, and terrorist propaganda) to more subtle ones, such as abusive language and spam, which are not always illegal, but nevertheless harmful. As harmful content is frequent online --- e.g.,~when surveyed for a 2019 report by the UK government, 23\% of the 12--15-year-olds stated to have observed it within the last year \cite{barker2019online} --- it is of particular concern to online communities, governments, and social media platforms.
With this in mind, we present the \arnav{first computing survey that} relates computational solutions to harmful content detection to online platform policies, with focus on analysing to what extent existing research efforts are suitable to address the types of harms online platforms aim to curb.

While combating harm is a high priority, preserving individuals' rights to free expression is also vital, which makes content moderation particularly difficult; yet, some form of moderation is clearly needed. Platform providers face very challenging technical and logistical problems in limiting harmful content, while at the same time wanting to allow a level of free speech that would enable rich and productive online interactions. %
Aiming to strike the right balance, many platforms institute guidelines and policies to specify what content is considered inappropriate. As manual filtering is hard to scale, and can even cause post-traumatic stress disorder in human annotators~\citep{mod_mental_health_fb},
there has been \arnav{significant} research effort to develop tools and technologies to automate \arnav{or assist in} the process.

\arnav{There have been several surveys of computational methods to detect and to address online harm.}
 \citet{schmidt2017survey} and \citet{fortuna2018survey} surveyed automated hate speech detection methods, but focused primarily on the features shown to be most effective in classification systems. \arnav{\citet{hardalov-etal-2022-survey} and \citet{guo-etal-2022-survey} surveyed automated methods 
to address mis- and disinformation,} \citet{salawu2020survey} provided an extensive survey on detecting cyberbullying, and \citet{vidgen2020directions} worked on cataloging abusive language training data. %

\arnav{There are some studies closely related to ours, that focus on the content guidelines of online platforms. \citet{gillespie-chap3-custodians} qualitatively studied several policies in the content guidelines of more than 60 platforms. They stated how content guidelines are the platforms' ``\textit{most deliberate and carefully crafted statement of principles}'' outlining what content is not permitted on the platform and why. They also suggested to look at inspecting guidelines of a variety of platforms in order to understand the overlap and the differences between them, as we do in this study. \citet{jiang-et-al-2020-guidelines}  analysed the relative focus that platforms place on types of abuse to moderate. They iteratively coded and analysed the community guidelines of eleven major social media platforms for coverage of 66 different types of policies. Through their analysis, they outlined the imbalance in coverage of policies, speculating that \textit{``platforms may have chosen to focus on and made rules regarding the types of misbehaviour that is most rampant on their platform, or made explicit the rules that are most reflective of their values''}. Our work builds on their study by categorising and computationally analysing the content guidelines of a larger set of platforms.}

\arnav{More concretely, we aim to bridge the gap between work on harmful content detection and platform solution requirements by surveying online platforms' content policies, and relating them to proposed approaches by the research community. 
We further quantify the extent of this disconnect, and identify under-explored directions by analysing the platform requirements based on the keyword usage in their T\&C clauses, and comparing those to what is currently available in the literature (both in published papers and in preprints). 
Nonetheless, no prior work 
has juxtaposed these policies with research directions and assessed the alignment between what platforms need vs. what technology has to offer. Hereby, we believe that such a survey is urgently needed.}

\section{Requirements of Online Platforms}\label{sec:platforms}

\emph{Harmful content} on online platforms can take on different forms. A 2019 white paper on online harms by the UK government considered the following sub-problems: online bullying and abuse, videos and images of children suffering sexual abuse, cyber-flashing, ``pile-on'' harassment, propaganda by terrorist groups, disinformation and misinformation \cite{barker2019online}. The paper noted that some of these are illegal, whereas others are lawful but potentially harmful.

\emph{Online platforms} is a broad term, representing various categories of digital content providers such as social media, online marketplaces, online video communities, dating websites and apps, support communities and forums, and online gaming communities. Each of them is governed by its own content moderation policies and follows its own definition of harmful content.
Below, we explore how online platforms define harmful content, and what content moderation policies they put in place accordingly by analysing their Terms and Conditions.
Even though these T\&Cs change over time, as platforms adapt to emerging issues such as the COVID-19 infodemic~\citep{alam-etal-2021-fighting-covid,10.1007/978-3-030-99739-7_52}, we %
analyse policies that we downloaded at a specific timestamp.
\footnote{Our offline Terms and Conditions archive are added in the Supplementary Material}

Table~\ref{tab:online_platforms} lists the 42 online platforms that we study. \arnav{We perform this selection with diversity of domain, content moderation needs of the platform as well as diversity of its user-base in mind in order to have a representative sample. Moreover, we only include platforms with publicly available content guidelines. Guidelines for what is acceptable or not on a platform is included either as part of their T\&Cs document or as a separate community guidelines document. As such, for our analysis, we choose the document based on where the platform lists this information.}
Note that we exclude ``alternative'' platforms such as \emph{Parler}, \emph{Bitchute}, \emph{Gab}, and \emph{Koo}, which are often specifically designed and commonly used as a substitute to platforms like \emph{Twitter} in order to circumvent their moderation;
these are already covered in designated studies~\citep{Buckley2022Censorship}. %
We categorise the 42 platforms based on the domain in which they operate, since the domain defines the type of content they need to moderate and is thus reflected in the respective T\&Cs. For instance, \textit{Reddit} is a \emph{generic forum} where any topic can be discussed, and thus it needs a wider range of policies compared to a \emph{specific forum} such as \emph{Motley Fool}, where the focus is on investing. The \emph{Mixed} category includes platforms that provide services (and hence, T\&Cs spanning multiple categories, e.g.,~\emph{Google} has policies that spread across its different services including \emph{Meet}, \emph{Mail}, \emph{Help Communities}, etc., each of which requires different expansiveness in its policy coverage. 

\begin{table}[t]%
\centering
\small
\begin{tabular}{cl}
\toprule
\textbf{Category} & \textbf{Platforms} \\ 
\midrule
{Dating} & Bumble \\ 
{Generic Forum} & \makecell{Quora, Reddit, Disqus, BG Mamma,\\ Discord, Something Awful, Substack, Clubhouse} \\
{Specific forum - Gaming} & Twitch, OverClocker UK\\
{Specific forum - Finance} & \makecell{Invstr, Money Saving Expert, Finimize, Public, \\StockTwits, Bogleheads,  Gastby, Motley Fool}\\
{Specific forum - Health} & \makecell{Mumsnet, Student Doctor Network, Patient, Doctissimo, Flo Health, Strava}\\
{Specific forum - Other} & Fiveable, Airbnb, Blind,  The Student Room, Shutterstock \\
{Online Marketplace} & Amazon, Depop, NTWRK, Rarible \\ 
{Social Media} & Facebook, YouTube, Twitter, Girl Tribe, TikTok \\ 
{Mixed} & Google, Spotify, Apple \\
\bottomrule
\end{tabular}
\caption{Classification of the 42 online platforms we study.}
\label{tab:online_platforms}
\end{table}

\subsection{Qualitative Analysis of the Terms and Conditions of Big Tech Platforms}

We a conducted qualitative analysis of the terms and conditions of big \emph{social media} and \emph{mixed} platforms, commonly known as \emph{Big Tech}.
Big Tech has stringent content moderation policies as well as the most advanced technology to detect harmful content. We summarise this analysis in Table~\ref{tab:policies}: we observe a lot of overlap, but also many differences.
\begin{table*}[tbh] 
\centering
    \resizebox{1.00\textwidth}{!}{%
\begin{tabular}{lccccc}
\toprule
\bf{Policy Clause} & \bf{Facebook$^\ddagger$} & \bf{Twitter} & \bf{Google} & \bf{Apple} & \bf{Amazon} \\ 
\midrule
\emph{Violence} & \checkmark & \checkmark & \checkmark & \implicit Intimidating, Threatening & \implicit Threatening \\ 
\emph{Dangerous} orgs/people & \checkmark & \checkmark & Maps, Gmail, Meet$^*$ & \vague Illegal act & \vague under illegal \\ 
\emph{Glorifying crime} & \checkmark & \checkmark & Maps, Gmail, Meet$^*$ & \vague Illegal act & \vague under illegal \\ 
\emph{Illegal goods} & \checkmark & \checkmark & Maps, Google Chat and Hangout, Drive, Meet$^*$ & \vague Illegal act & \vague under illegal \\ 
\emph{Self-harm} & \checkmark & \checkmark & \checkmark & \cross & \cross \\ 
\emph{Child sexual abuse} & \checkmark & \checkmark & \checkmark & \vague & \cross \\ 
\emph{Sexual Abuse (Adults)} & \checkmark & \checkmark & \cross & \vague & \vague \\ 
\emph{Animal abuse} & \checkmark & \vague Sensitive media policy & Earth, Drive, Meet$^*$ & \vague Illegal act & \vague under illegal \\ 
\emph{Human trafficking} & \checkmark & \checkmark & \checkmark & \vague Illegal act & \vague under illegal \\ 
\emph{Bullying and harassment} & \checkmark & \checkmark & \checkmark & \checkmark & \implicit Threatening \\ 
\emph{Revenge porn} & \checkmark & \checkmark & \cross & \cross & \vague obscene \\ 
\emph{Hate Speech} & \checkmark & \checkmark Hateful Conduct & \checkmark & \checkmark & \implicit Threatening \\ 
\emph{Graphic content} & \checkmark & \checkmark & Maps$^*$ & \cross & \vague \\ 
\emph{Nudity and pornography} & \checkmark & \checkmark & Earth, Meet, Drive, Chat and Hangout$^*$ & \checkmark & \vague under obscene \\ 
\emph{Sexual Solicitation} & \checkmark & \cross & Maps$^*$ & \cross & \cross \\ \emph{Spam} & \checkmark & \checkmark & \checkmark & \cross & \checkmark \\ 
\emph{Impersonation} & \checkmark & \checkmark & Maps, Earth, Chat and Hangout, Gmail, Meet$^*$ & \checkmark & \checkmark \\ 
\emph{Misinformation} & \checkmark False news & \checkmark & Maps, Drive$^*$ & \checkmark & \cross \\ 
\emph{Medical Advice} & \vague & COVID-19 specific & Drive$^*$ & \cross & \cross \\ 
\bottomrule
\end{tabular}
}
\caption{Summary of the terms and conditions of Big Tech. We use the following notation:
{\protect\checkmark} -- explicitly mentioned in the policy;
{\cross} -- not mentioned in the policy;
{\implicit} -- implicitly mentioned in the policy;
{\vague} -- broadly mentioned in the policy under a more generic clause;
$^*$ mentioned in additional service-specific policy;
$^\ddagger$ -- the same policy applies to Instagram.
}
\label{tab:policies}
\end{table*}

\emph{Facebook} covers most policy clauses related to harmful language, though, e.g.,~the coverage of \emph{medical advice} is described quite broadly, and 
it is thus unclear whether users are free to share medical advice in their posts.
\emph{Google}'s terms of service cover basic guidelines on acceptable conduct. The more specific clauses regarding \emph{hate speech, bullying}, and \emph{harassment} are in service-specific policies, as shown in Table~\ref{tab:policies}.
\emph{Amazon} and \emph{Apple} offer very different services compared to \emph{Facebook}, \emph{Twitter}, and \emph{Google}, which is reflected in the clauses covered in their terms of service.

\emph{Apple}'s policies very broadly mention %
\emph{dangerous organisations, glorifying crime, illegal goods, child sexual abuse, sexual abuse (adults), animal abuse}, and \emph{human trafficking} under \emph{illegal acts}. \emph{Violence} falls under threatening and intimidating posts. Their policy does not mention \emph{medical advice, spam, sexual solicitation, revenge porn, graphic content}, and \emph{self-harm}. 
For \emph{Amazon}, \emph{dangerous organisations and people, glorifying crime, illegal goods, child sexual abuse, sexual abuse (adults), animal abuse}, and \emph{human trafficking} are
mentioned under \emph{illegal} acts, which leaves room for ambiguity. \emph{Revenge porn, graphic content, nudity and pornography} are also broadly mentioned in clauses pertaining to obscenity. However, there are no policies in place that directly address \emph{medical advice, misinformation, sexual solicitation}, and \emph{self-harm}.

\subsection{Quantitative Analysis of all platforms' terms and conditions}%

In order to understand to what degree these patterns exist more widely, we further conducted a quantitative analysis of the T\&Cs across the 42 platforms that we consider in our study (see Table~\ref{tab:online_platforms}).

\subsubsection{Methodology}

We quantified the presence of specific categories in the T\&Cs by searching for policy-related keywords and by counting the number of occurrences we found. We compiled the keyword lists semi-automatically %
by looking for words and phrases commonly mentioned in the T\&Cs when discussing a given topic. For instance, the keyword list for \emph{hate speech} includes \emph{racist}, \emph{ageist}, \emph{disablist}, and \emph{homophobic}, among others. A caveat to note here is that the number of search keywords differs across the policy clauses, which could affect the number of matches we found. %
Moreover, the T\&Cs are often long, wide-ranging, and spread across several documents. Thus, in order to perform a more comprehensive search, we scraped all pages hyperlinked in the main T\&C document. We further converted them to lowercase and we removed emails, numbers, URLs, and punctuation.

\subsubsection{Results}

\paragraph{By platform category}
We outline the policy presence across all platforms in Figure~\ref{fig:plat_cat}. Confirming what we observed in the qualitative analysis, 
we found that social media platforms tended to have the most expansive policies, with all topics being present in the T\&Cs of \emph{Facebook} and \emph{Twitter}, while \emph{TikTok} and \emph{YouTube} covered all but \emph{sexual solicitation}.

The outlier here is \emph{Girl Tribe by MissMalini}, which is a much smaller platform exclusively for women. The expansiveness compared to other categories could be attributed to the additional level of public scrutiny that social media companies are subject to.

While on average, \emph{generic fora} cover more categories compared to \emph{specific fora}, there is considerable variance within the specific fora with \textit{Gaming} having a higher average policy coverage compared to \textit{generic fora}, whereas fora pertaining to \textit{Finance} have the least coverage among all categories. We attribute this to the increased attention that \emph{Gaming} fora have received in the media as well as by regulatory authorities, and to accusations that they are not taking enough action to moderate harmful content~\citep{bbc_twitch_2021}.

\begin{figure}[t]
    \centering
    \includegraphics[width=0.9\columnwidth]{plots/policy_cat.png}
    \caption{Number of policy topics covered by each platform per category.}
    \label{fig:plat_cat}
\end{figure}

\begin{figure}[h]
     \centering
     \begin{subfigure}[b]{0.48\textwidth}
         \centering
         \includegraphics[width=\textwidth]{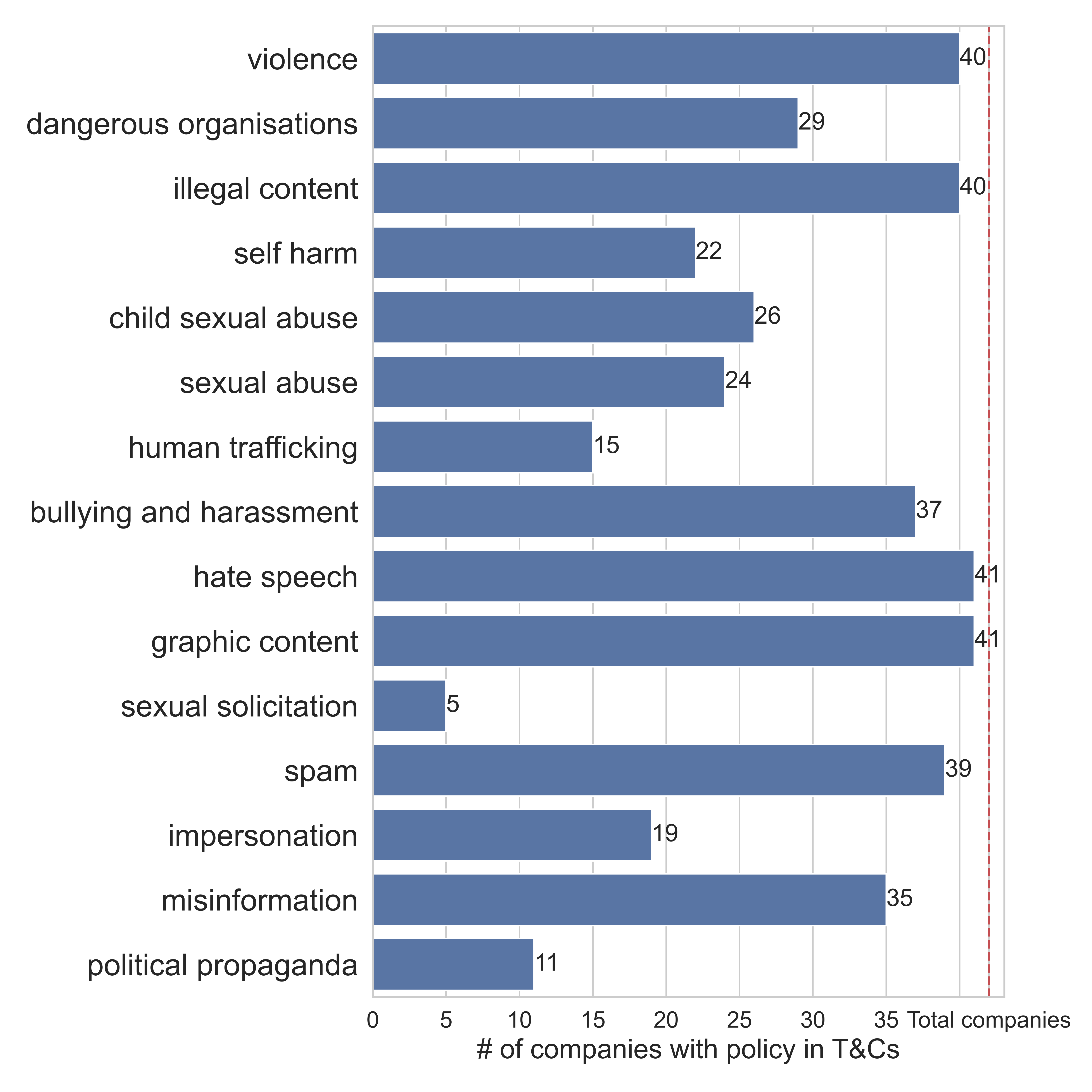}
         \caption{number of platform T\&Cs out of 42 mentioning each topic}
         \label{fig:policy_company}
     \end{subfigure}
     \hfill
     \begin{subfigure}[b]{0.48\textwidth}
         \centering
         \includegraphics[width=\textwidth]{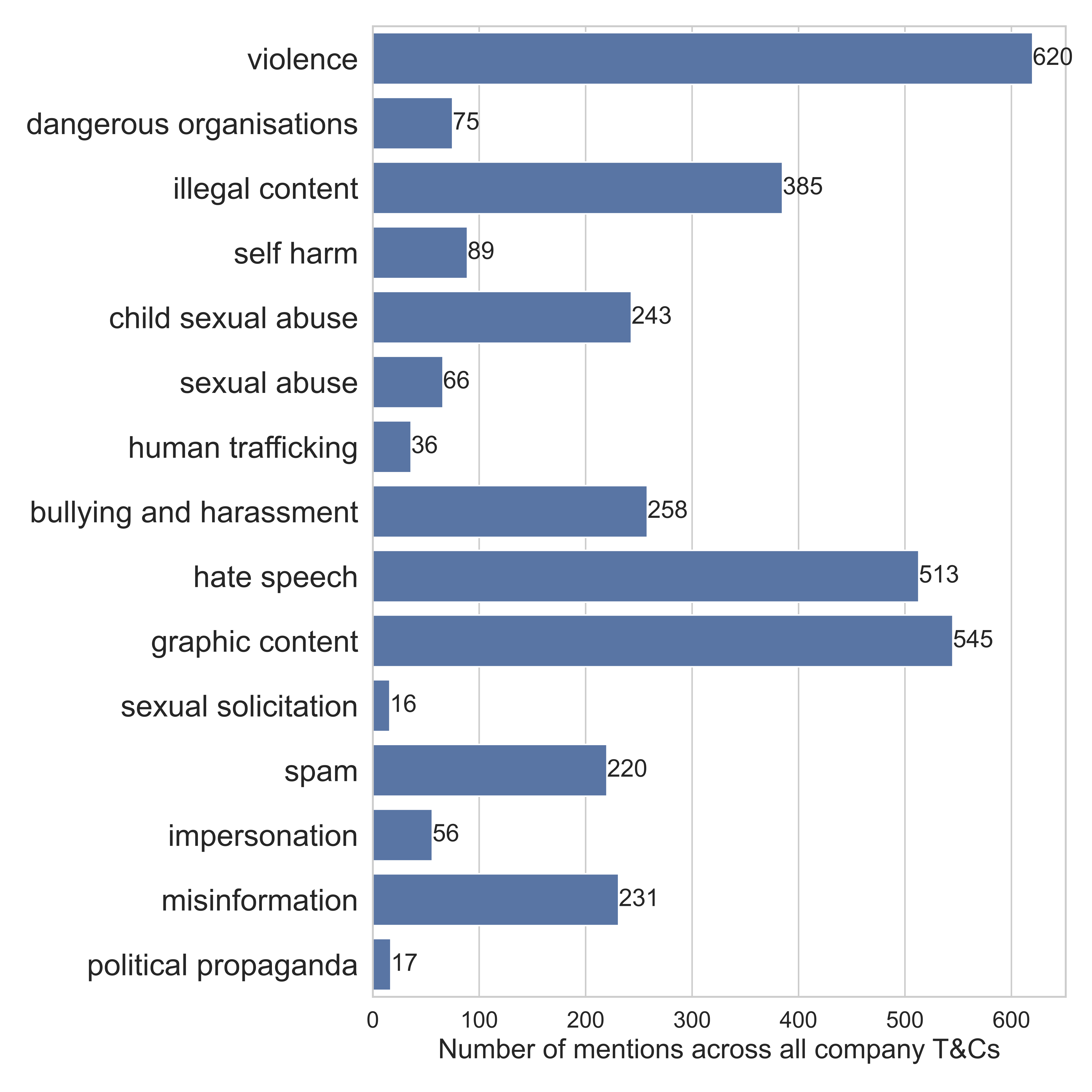}
         \caption{number of mentions of each topic across all platform T\&Cs}
         \label{fig:policy_absolute}
     \end{subfigure}
     \caption{Topic mentions across the platform Terms and Conditions (T\&Cs).}
\end{figure}

\paragraph{By policy}
Interesting patterns can be observed in a per-policy analysis of the T\&C documents. Figure~\ref{fig:policy_company} plots the number of platforms that cover each policy, which gives a sense of relative requirement for moderation for these policies across the platforms. We can see that \emph{Hate speech} and \emph{graphic content} policies are included in the T\&Cs of all platforms except for \emph{Gatsby}\footnote{Gatsby does not cover any of the policies taken into account in this work}, followed by \emph{violence}, \emph{illegal content} and \emph{spam}, respectively.
\emph{Sexual Solicitation} is the least covered, with only some \textit{Social Media} and \textit{Gaming fora} \arnav{listing it.\footnote{Yet, this kind of content is illegal in certain parts of the world and can have wide-ranging consequences; many platforms had to scramble to remove sexual solicitation and sexual content from their websites, most notably Craigslist removed their entire personals section to deal with the issue.}} \emph{Self-harm} is covered by all \textit{Social Media} platforms, but it is not explicitly stated in half of the \textit{Health} fora and in the \textit{Dating} platform \emph{Bumble}'s policies, where it ought to be important. A similar pattern can be observed for \emph{Sexual Abuse} and \emph{Human Trafficking}, which have low coverage in \textit{Dating} and \textit{Health} fora.

Figure~\ref{fig:policy_absolute} shows the number of times each topic is mentioned across all platform T\&Cs. While multiple mentions do not perfectly capture how important a particular topic is, it does give us a sense of how much focus a platform is paying to a particular policy clause across several spread out pages of its T\&Cs. Similarly to Figure~\ref{fig:policy_company}, \emph{Violence, Graphic Content} and \emph{Hate Speech} are the most mentioned topics, demonstrating the significant attention paid to them as well as the importance of detecting policy violations regarding these topics. \emph{Sexual Solicitation} and \emph{Political Propaganda} are the topics with the fewest mentions. For \emph{Political Propaganda}, this can be partly attributed to its narrower focus, e.g.,~\emph{Gaming} and \emph{Finance} fora are unlikely to host large amounts of political content. \emph{Child Sexual Abuse} is in the same range of mentions as \emph{Misinformation, Bullying and Harassment} and \emph{Spam}, while being covered by fewer platforms.

\section{Automatic Harmful Content Detection}
\label{sec:harm}

Below, we first analyse the publications in arXiv and to what extent they cover the above-described topics of concern in the T\&Cs, and then we discuss some specific research, with focus on offensive language.

\subsection{Analysing Publications in arXiv}
\label{subsec:arxiv_pubs}

In order to estimate how much research attention has been paid to each T\&C topic, we looked for mentions of policy-related keywords in arXiv\footnote{https://arxiv.org/} within the \texttt{cs} subject classification for {2015--2021}, 
taking the number of results as a proxy for the amount of work conducted pertaining to that topic. We show the results in Figure~\ref{fig:arxiv_all}. 
We can see that the number of papers on \emph{Illegal Content} is consistently higher than for the other topics. We further see that work on \emph{Hate Speech} and \emph{Misinformation} has seen a steep growth since 2017, which reflects the attention these topics have attracted in recent years.
\emph{Spam} is another topic with a consistently high volume of publications. In contrast, topics such as \emph{Sexual Solicitation}, \emph{Child Sexual Abuse}, and \emph{Human Trafficking} have seen very little research. We attribute this to the difficulty of getting access to datasets and performing research, due to the sensitive nature of these topics.

\begin{figure}[h]
    \centering
    \includegraphics[width=\textwidth]{plots/arxiv_temporal_all.png}
    \caption{Number of papers found in arXiv for the different topics over time.}
    \label{fig:arxiv_all}
\end{figure}

Work on harmful content detection has addressed several aspects of the above issues, but we are unaware of any work that covers them comprehensively. \citet{vidgen2020directions} attempted to categorise the available datasets with focus on abusive language detection, analysing a total of 64 datasets.
They found that around half of the datasets covered English only, with some datasets covering European languages, Hindi and Indonesian, and six datasets covering Arabic. The primary source of the data was Twitter, but there were also datasets using content from other social networks. Overall, the size of most datasets was under 50,000 instances, and under 5,000 instances for half of the datasets.

\arnav{
\subsection{Approaches}}
\label{subsec:approaches}

There has been a lot of research effort aimed at detecting specific types of offensive content, e.g.,~hate speech, offensive language, cyberbullying, and cyber-aggression. Below, we briefly describe some relevant tasks, datasets, and approaches.

\paragraph{Hate Speech Detection} This is by far the most studied harmful language detection task~\cite{waseem-hovy-2016-hateful, kwok2013locate,burnap2015cyber,ousidhoum-etal-2019-multilingual,chung-etal-2019-conan,sheikh_domain_adaptation}. %
A recent shared task on the topic is HateEval~\cite{basile2019semeval} for English and Spanish.
The problem was also studied from a multimodal perspective, e.g.,~\citet{DBLP:journals/corr/abs-1910-02334} developed a collection of 5,020 memes for hate speech detection. More recently, the Hateful Memes Challenge by Facebook introduced a dataset consisting of more than 10K memes, annotated as hateful or non-hateful~\cite{kiela2020hateful}: the memes were generated {\em artificially}, so that they resemble real memes shared on social media, along with ``benign confounders.''
More recent work studied \emph{harmful memes} \citep{pramanick-etal-2021-detecting,pramanick-etal-2021-momenta-multimodal,DISARM} and propagandistic memes~\citep{dimitrov-etal-2021-detecting,dimitrov-etal-2021-semeval}.

\paragraph{Offensive Language Detection} One of the most widely used datasets is the one by~\citet{davidson2017automated}, which contains over 24,000 English tweets labelled as \emph{non-offensive}, \emph{hate speech}, and \emph{profanity}. There have been several shared tasks with associated datasets that focused specifically on offensive language identification, often featuring multiple languages: OffensEval 2019--2020~\cite{zampieri-etal-2019-semeval,zampieri-etal-2020-semeval} for English, Arabic, Danish, Greek, and Turkish, GermEval~2018~\cite{wiegand2018overview} for German, HASOC~2019~\cite{mandl2019overview} for English, German, and Hindi, TRAC 2018--2020 for English, Bengali, and Hindi~\cite{fortuna2018merging,trac2020}.
Offensive language was also studied from a multimodal perspective, e.g.,~\citet{mittos2019and} analysed memes shared on 4chan as \emph{offensive} vs. \emph{non-offensive}.

\paragraph{Aggression Detection} The TRAC shared task on Aggression Identification~\cite{kumar2018benchmarking} provided participants with a dataset containing 15,000 annotated Facebook posts and comments in English and Hindi for training and validation. For testing, two different sets, one from Facebook and one from Twitter, were used. The goal was to discriminate between three classes: \emph{non-aggressive}, \emph{covertly aggressive}, and \emph{overtly aggressive}. 

\paragraph{Toxic Comment Detection} The Toxic Comment Classification Challenge~\cite{jigsaw}
was a Kaggle competition, which provided participants with almost 160K comments from Wikipedia organised in six classes: \emph{toxic, severe toxic, obscene, threat, insult}, and \emph{identity hate}. The dataset was also used outside of the competition~\cite{georgakopoulos2018convolutional}, including as additional training material for the TRAC shared task~\cite{fortuna2018merging}. It task was later extended to multiple languages~\cite{jigsaw-multilingual},\footnote{\href{http://www.kaggle.com/c/jigsaw-multilingual-toxic-comment-classification}{\href{http://www.kaggle.com/c/jigsaw-multilingual-toxic-comment-classification}https://www.kaggle.com/c/jigsaw-multilingual-toxic-comment-classification}}
offering 8,000 Italian, Spanish, and Turkish comments.  Recently, \citet{juuti-etal-2020-little} presented a systematic study of data augmentation techniques in combination with state-of-the-art pre-trained Transformer models for toxic language detection.
A related Kaggle challenge features Detecting Insults in Social Commentary.\footnote{\url{http://www.kaggle.com/c/detecting-insults-in-social-commentary}}
Other datasets include Wikipedia Detox~\cite{wulczyn2017ex},
and the dataset by \citet{davidson2017automated}.

\paragraph{Cyberbullying Detection}  \citet{xu2012learning} used sentiment analysis and topic models to identify relevant topics, \citet{dadvar2013improving} relied on user-related features such as the frequency of profanity in previous messages, and \citet{ROSA2019333} presented a systematic review of automatic cyberbullying detection research.

\paragraph{Abusive Language Detection} 
 There have also been datasets that cover various types of abusive language. \citet{founta2018large} tackled hate and abusive speech on Twitter, introducing a dataset of 100K tweets. \citet{glavas-etal-2020-xhate} targeted hate speech, aggression, and attacks in three different domains: Fox News (from GAO), Twitter/Facebook (from TRAC), and Wikipedia (from WUL). In addition to English, it further offered parallel examples in Albanian, Croatian, German, Russian, and Turkish. However, the dataset was small, containing only 999 examples. Among the popular approaches for abusive language detection are cross-lingual embeddings~\cite{ranasinghe-zampieri-2020-multilingual}, cross-lingual neighbourhood transformer representations~\cite{sheikh_cross_lingual}, and deep learning~\cite{founta2019unified}.
As abusive language has many aspects, there have been several proposals for multi-level taxonomies covering different types of abuse. This enables understanding how different types and targets of abusive language relate to each other, and informs efforts to detect them. For example, offensive messages targeting a group are likely \emph{hate speech}, while when the target is an individual, this is likely \emph{cyberbullying}.
\citet{waseem2017understanding} presented a taxonomy that differentiates between (abusive) language directed towards a specific individual or entity, or towards a generalised group, as well as between explicit and implicit abusive content. \citet{wiegand2018overview} extended this idea to German tweets: they developed a model to detect offensive vs.~non-offensive tweets, and further sub-classified the former as profanity, insult, or abuse. \citet{zampieri-etal-2019-predicting} developed a very popular three-level taxonomy, which considers both the type and the target of offence. 
The taxonomy served as the basis of the \olid{} dataset of 14,000 English tweets, which was used in two shared tasks (OffensEval) at SemEval in 2019~\cite{zampieri-etal-2019-semeval} and in 2020~\cite{zampieri-etal-2020-semeval}. For the latter, an additional large-scale dataset was developed, consisting of nine million English tweets labelled in a semi-supervised fashion~\cite{SOLID}. This new dataset enabled sizeable performance gains, especially at the lower levels of the taxonomy.
The taxonomy was also adopted for Arabic~\cite{mubarak-etal-2021-arabic}, Danish \cite{sigurbergsson2020offensive}, Greek \cite{pitenis2020}, and Turkish \cite{coltekin2020}.

\emph{Troll detection} has been addressed using semantic analysis~\cite{cambria2010not}, domain-adapted sentiment analysis~\cite{Seah2015}, various lexico-syntactic features about user writing style and structure~\cite{chen2012detecting,mihaylov-etal-2015-finding,mihaylov-nakov-2016-hunting},
as well as using graph-based approaches~\cite{Ortega20122884,kumar2014accurately}.
There have also been studies on general troll behaviour~\cite{herring2002searching,buckels2014trolls,InternetResearchJournal:2018,nakov-etal-2017-trust} and role~\citep{mihaylov-etal-2015-exposing,atanasov-etal-2019-predicting}, cyberbullying~\cite{galan2015supervised,sarna2015content,WONG2018247,DBLP:journals/intr/SezerYY15}, 
as well as on linking fake troll profiles to real users \cite{galan2015supervised}. 
Some studies related to cyberbullying have already been applied in real settings in order to detect and to stop cyberbullying in elementary schools using a supervised machine learning algorithm that links fake profiles to real ones on the same social medium~\cite{galan2015supervised}. 

\emph{Identification of malicious accounts} in social networks is another important research direction. This includes detecting spam accounts~\cite{almaatouq2016if,mccord2011spam}, fake accounts~\cite{fire2014friend,cresci2015fame}, compromised accounts and phishing accounts~\cite{adewole2017malicious}. Fake profile detection has also been studied in the context of cyberbullying~\cite{galan2015supervised}. 

\emph{Web spam detection} was addressed text classification \cite{sebastiani2002machine}, e.g.,~using spam keyword spotting~\cite{dave2003mining}, lexical affinity of words to spam content~\cite{hu2004mining}, frequency of punctuation, and word co-occurrence~\cite{li2006combining}. 

\paragraph{\arnav{Methods}}
\label{sec:methods}
The most common way to address the above tasks is to use pre-trained transformers~\citep{vaswani2017attention}: typically BERT~\cite{devlin2019bert}, but also RoBERTa~\cite{liu2019roberta}, ALBERT~\cite{lan2019albert}, and GPT-2~\cite{radford2019language}. In a multi-lingual setup, also mBERT~\cite{devlin2019bert} and XLM-RoBERTa~\cite{conneau2019unsupervised} have proved useful.
CNNs~\cite{fukushima1980neocognitron}, RNNs~\cite{rumelhart1986learning}, and GRUs~\cite{cho2014learning}, including ELMo~\cite{peters:2018}. 
Older models such as SVMs
are sometimes also used, typically as part of ensembles. Moreover, lexica such as HurtLex~\cite{DBLP:conf/clic-it/BassignanaBP18} and Hatebase\footnote{\url{http://hatebase.org/}} are also popular.
\arnav{For multi-modal detection of these harms, }popular approaches include Visual BERT, ViLBERT, VLP, UNITER, LXMERT, VILLA, ERNIE-Vil, Oscar and various Transformers \citep{li2019visualbert,su2020vlbert,zhou2020unified,tan2019lxmert,gan2020largescale,yu2020ernievil,li2020oscar,lippe2020multimodal,zhu2020enhance,muennighoff2020vilio,zhang2020hateful,kumar-nandakumar-2022-hate}.

\section{Lessons Learned and Major Challenges}\label{sec:lessons}

Despite the stringent policies in place, the amount of harmful content online keeps growing. Research is complicated as online platforms have different understanding of harmful content, and platform's Terms and Conditions (T\&Cs) try to imagine every possible outcome of particular statements, while trying to be transparent about the areas they fall short in. The reality is that some clauses in the T\&Cs are hard to automate and require human attention and understanding.
Below, we describe some of these challenges in more detail. The list is not exhaustive, but it illustrates the main challenges faced by online platforms and researchers, as new issues regularly appear, policies evolve, and new detection technologies are constantly being developed.

\paragraph{Mismatch in Goals}
Table~\ref{tab:comparison} shows a comparison of the number of papers on arXiv that mention a specific type of harmful content and the corresponding number of platforms that mention that type in their T\&Cs. We can see that \emph{graphic content} and \emph{violence} are relatively less researched in spite of being covered by large number of platforms. On the other hand, \emph{misinformation, illegal content}, etc. have attracted a lot of research. 
As is readily apparent when comparing 
T\&Cs documents of online platforms (Section~\ref{sec:platforms}) to harmful content tasks tackled in research (Section~\ref{sec:harm}),
there is a clear disconnect between what types of content platforms seek to moderate, and what current harmful content detection research offers, as demonstrated by the high variance in the ratio column in Table~\ref{tab:comparison}.

\paragraph{Lack of participatory design.}
While this work focuses on platform policies and academic literature around content moderation, it is the users of a platform who are at the receiving end of decisions based on these policies and algorithms. Hence, a people-centric and participatory process model for AI-assisted content moderation is crucial~\citep{ethical_scaling_2022}. Following a one-size-fits-all approach does not work when norms, values, and what is considered harmful differs across communities~\citep{chandrasekharan-et-al-2018-norms} and cultures~\citep{jiang-et-al-2021-perceptions}.

\paragraph{Policy challenges.} It is hard to define harmful content policy that is congruent with the harm that such content inflicts. For example, Facebook's hate speech policy is tiered, where Tier 1 aims to define the kind of hate speech that is considered most harmful. The policy covers content targeting a person or a group of people based on their protected characteristics such as race, ethnicity, national origin, disability, gender, etc. 
According to this policy, certain offensive statements are considered more harmful than others. \arnav{Moreover, a number of policies are left under-specified or are ambiguous in what content is allowed on the platform. The policies are also fluid, change based on the circumstances, and are affected by events in the real world~\citep{vengattil_facebook_2022}, which makes them challenging for researchers to follow.}

\paragraph{Context.} An obvious challenge that comes with the need to moderate \emph{unlawful} content is that platforms are operating in multiple countries with a wide spectrum of legal standards. Similarly, the same content considered in different contexts can be either perfectly benign or can be perceived as harmful. Such context also changes with time, e.g.,~certain slur words are appropriated by their target groups as reclaimed speech. Such historical, legal, and cultural context is not explicitly encoded into machine learning models at present \arnav{and makes it impossible for them to accurately detect and for platforms to moderate this content.}
\paragraph{Dataset Availability} 
As outlined earlier, data for sensitive policies such as \emph{sexual solicitation} and \emph{child sexual abuse} are hard to collect and share. However, even for categories such as \emph{misinformation} and \emph{bullying and harassment}, access to data is primarily at the discretion of the platforms. While there have been efforts to increase research access, these remain insufficient. Moreover, since violating content is often taken down, researchers \arnav{enter a timed race with platforms in their efforts to collect content before it is removed or only have access to the content that platforms do not remove, making it incredibly} hard for them to develop systems capable of detecting or assessing the potential harms caused by such content. Hence, there is a need for platforms 
to develop privacy-preserving ways of sharing datasets.

\begin{table*}[t] 
\centering
\resizebox{0.6\textwidth}{!}{%
\begin{tabular}{lrcr}
\toprule
{\bf Topic} & \bf arXiv Papers & \bf Online Platforms &  \bf Ratio \\
\midrule
\centering
sexual solicitation     &           1 &                 5 &   0.20 \\
child sexual abuse      &          12 &                26 &   0.46 \\
sexual abuse            &          24 &                24 &   1.00 \\
graphic content         &          79 &                41 &   1.93 \\
human trafficking       &          53 &                15 &   3.53 \\
self harm               &         102 &                22 &   4.64 \\
dangerous organisations &         142 &                29 &   4.90 \\
violence                &         255 &                40 &   6.38 \\
bullying and harassment &         276 &                37 &   7.46 \\
impersonation           &         192 &                19 &  10.11 \\
spam                    &         434 &                39 &  11.13 \\
political propaganda    &         209 &                11 &  19.00 \\
hate speech             &        1089 &                41 &  26.56 \\
illegal content         &        1113 &                40 &  27.82 \\
misinformation          &        1029 &                35 &  29.40 \\
\bottomrule
\end{tabular}}
\caption{Relative research focus in terms of number of research papers in arXiv that study a given type of harmful content and the number of online platforms that aim to moderate such content, as well as a ratio thereof.}
\label{tab:comparison}
\end{table*}

\paragraph{Content is Interlinked.} People use text, images, videos, audio files, GIFs, emojis --- all together --- to communicate. While each modality by itself may be seemingly benign, when combined, they might make content that violates the T\&Cs. Thus, there is a need for holistic understanding of the content.

\paragraph{Dataset Specificity} Most publicly available datasets are specific to the type of harm, to the targeted group, or to other fine-grained distinctions, as harmful content can be highly context-sensitive. \arnav{Further, in the hierarchical annotation schemes introduced by these datasets, lower levels of the taxonomy contain a subset of the instances in the higher levels, and thus there are fewer instances in the categories in each subsequent level.} This diversity has led to the creation of many datasets, most of which are small in size, and much of the research on harmful content detection is fragmented, with very few studies viewing the problem from a holistic perspective.

\section{Conclusion}

We discussed recent methods for harmful content detection in the context of the content policies of online platforms. \arnav{We inspected the content guidelines of 42 platforms and compared the needs highlighted there to available research.}
\arnav{We found a substantial imbalance in the relative focus to different policy clauses in research vs. by platforms. For instance, misinformation and political propaganda are among the clauses with a high research-papers-to-platform-coverage ratio, while sexual solicitation and graphic content received limited attention, despite being critical for a number of platforms. While academic research should not aim to follow industrial needs, shedding light on this relative focus can aid in identifying major challenges and possible remedies that harmful content detection faces going forward. We argued that current research does not produce the tools for automating content moderation on online platforms that matched their needs. We further outlined some major challenges that need to be addressed.}

\bibliography{iclr2023_conference}
\bibliographystyle{iclr2023_conference}

\end{document}